\documentclass[lettersize,journal]{IEEEtran}

\usepackage{amsmath,amsfonts}
\usepackage{algorithmic}
\usepackage{algorithm}
\usepackage{array}
\usepackage[caption=false,font=normalsize,labelfont=sf,textfont=sf]{subfig}
\usepackage{textcomp}
\usepackage{stfloats}
\usepackage{url}
\usepackage{verbatim}
\usepackage{graphicx}
\usepackage{cite}
\usepackage{xcolor}

\ifodd 0
\newcommand{\rev}[1]{{\color{blue}#1}} 
\else
\newcommand{\rev}[1]{#1}
\fi

\begin{document}

\title{Pushing Large Language Models to the 6G Edge: Vision, Challenges, and Opportunities}
\author{Zheng Lin, Guanqiao Qu,  Qiyuan Chen, Xianhao Chen,~\IEEEmembership{Member,~IEEE},  Zhe Chen,~\IEEEmembership{Member,~IEEE}, and \\Kaibin Huang,~\IEEEmembership{Fellow,~IEEE}
\thanks{The work was supported in part by the Research Grants Council of Hong Kong under Grant 27213824 and in part by HKU IDS Research Seed Fund under Grant IDS-RSF2023-0012.}
\thanks{Zheng Lin, Guanqiao Qu, Qiyuan Chen, Xianhao Chen, and Kaibin Huang are with the Department of Electrical and Electronic Engineering, University of Hong Kong, Pok Fu Lam, Hong Kong SAR, China. Xianhao Chen is also with HKU Musketeers Foundation Institute of Data Science, University of Hong Kong, Pok Fu Lam, Hong Kong SAR, China. (e-mail: linzheng@eee.hku.hk; gqqu@eee.hku.hk; qiyuanchen@connect.hku.hk; xchen@eee.hku.hk; huangkb@eee.hku.hk).
Zhe Chen is with the School of Computer Science, Fudan University, Shanghai, China. (e-mail: zhechen@fudan.edu.cn).
\textit{(Corresponding author: Xianhao Chen)}}}

%
%

\markboth{}%
{Shell \MakeLowercase{\textit{et al.}}: A Sample Article Using IEEEtran.cls for IEEE Journals}


\maketitle

\begin{abstract}
Large language models (LLMs), which have shown remarkable capabilities, are revolutionizing artificial intelligence (AI) and shaping our society. However, the status quo cloud-based LLM deployment faces critical challenges such as long response time and privacy violation, while on-device LLM deployment is hindered by the limited capabilities of end devices.  To address these issues, this article explores the transformative potential of deploying LLMs at the 6G edge. We first introduce killer applications to exemplify the urgent need for edge LLM deployment. Meanwhile, we identify the inherent limitations of on-device LLM deployment. We therefore argue that \textit{end-edge cooperation} at the 6G edge is a promising solution for the dilemma. Towards this end, we elaborate on the 6G MEC architecture tailored for LLMs. Furthermore, we delve into edge training and edge inference for LLMs, with a focus on end-edge cooperation. In both aspects, we discuss a spectrum of cutting-edge techniques, including split learning/inference, parameter-efficient fine-tuning, parameter-sharing inference, and small-large language model cooperation. Finally, we investigate open problems in green and privacy-preserving edge LLM deployment. This work provides a comprehensive and forward-looking perspective and pathways for enabling LLM deployment at the network edge.
\end{abstract}
\vspace{-1mm}
\section{Introduction\label{introduction}}
\vspace{-1mm}
The advent of large language models (LLMs), powered by the success of transformer architectures, has revolutionized artificial intelligence (AI) and attracted global attention. Nowadays, major players in the AI industry are vying to develop their own LLMs, with notable examples including OpenAI's GPT-4, Google's PaLM 2, and Meta's LLaMA 2. These models, trained on vast and diverse datasets from the Internet, exhibit emergent generalization capabilities as their model size substantially increases -- a phenomenon referred to as ``emergence''. For instance, GPT-4, with its massive scale, can successfully perform tasks such as arithmetic reasoning or logic-based problem-solving, even without explicit training on those tasks~\cite{achiam2023gpt}. Such exceptional capabilities make LLMs highly versatile, enabling direct application or easy adaptation (e.g., fine-tuning or instruction tuning) to a broad spectrum of downstream tasks, thereby unlocking unprecedented potential in various applications, such as Chatbot, content generation, healthcare, and robotics.

Unfortunately, the existing LLM products predominantly rely on cloud computing, which suffers from excessive latency, high bandwidth cost, and severe privacy concerns. First, the cloud-based model inference is inadequate for real-time applications (e.g., LLM-empowered robotics control/navigation/exploration) due to long data transmission delay. Second, the emergence of multimodal LLMs requires input/output of not only texts, but also images, videos, audio, and other sensory data. Centralizing these massive data for training or inference will consume significant backhaul/backbone network bandwidth and overwhelm the central cloud infrastructure, which is not scalable. At last, LLM training or inference raises severe privacy concerns, particularly considering applications involving highly sensitive data, such as medical data or human activities including audio instructions and gestures at home. As a result, there is an urgent need to finetune and deploy LLMs on or in closer proximity to data sources while preserving data ownership of end users.

The growing trend of on-device LLM deployment appears to an effective solution for the above needs. However, while on-device LLMs mitigate the limitations of cloud LLMs by reducing data transmissions and privacy leakage, this paradigm is inherently hampered by the scarce computing capabilities of edge devices. Consequently, the popular on-device LLMs typically have fewer than 10 billion parameters, e.g., the Gemini Nano-2 model comprising 3.25 billion parameters (3GB for 32-bit floats), limiting their performance and application scope. Also, current implementations primarily focus on on-device LLM inference, neglecting another critical aspect: on-device training. This shortfall prevents models from being improved or personalized for individual user needs, ultimately leading to unsatisfactory performance.

As we are progressing towards the early standardization of 6G, it is widely recognized that 6G will evolve into a mobile network supporting in-network and distributed AI at the edge. Based on mobile edge computing (MEC), edge devices and MEC servers can \textit{cooperate} in LLM inference and training to mitigate the aforementioned dilemma, a paradigm we call ``end-edge cooperation'' in this paper. Specifically, compared with cloud LLMs, end-edge cooperation offers low transmission latency due to localized computing power close to end users;  Compared with on-device LLMs, end-edge cooperation provides significantly more computing capabilities than edge devices alone. From a technical perspective, split machine learning (including both split learning and inference) can serve as a cornerstone for end-edge cooperation by partitioning the intensive workload over distributed edge devices/servers while still preserving data privacy. Furthermore, other emerging techniques, such as speculative decoding, can also facilitate end-edge cooperation by running small models on edge devices and large models on edge servers, thereby satisfying accuracy and latency requirements simultaneously.


Motivated by above pressing needs, in this paper, we will explore the exciting research frontier at the intersection of LLM deployment and 6G MEC. Specifically, we first provide motivating killer MEC applications for pushing LLMs to the edge. Meanwhile, we demonstrate the inherent limitations of on-device LLM deployment, demonstrating that purely relying on-device capabilities may be insufficient. Towards this direction, we present the tailored 6G MEC architecture for LLM deployment, followed by the integrated communication-computing technologies for LLM training and inference from the perspective of end-edge cooperation. It is noted that unlike prior studies, such as~\cite{shen2023large}, discussing how to leverage LLMs to optimize wireless networks or MEC (i.e., \textit{LLMs for networks}), this article shifts the perspective to how to leverage 6G MEC to support LLM training and inference (i.e., \textit{networks for LLMs}). Furthermore, this article also differs from articles on on-device LLM deployment since end-edge cooperation is our focus.

The rest of this paper is organized as follows. Section~\ref{application}
introduces the killer applications. Section \ref{challenges} identifies the challenges, followed by an overview of MEC architecture tailored for LLMs in Section~\ref{sec:architecture}. End-edge training and inference for LLMs are discussed in Section~\ref{sec:fine_tuning} and Section~\ref{inference}, respectively.
Open problems are identified in Section~\ref{Open} and the conclusions are drawn in Section~\ref{conclusion}.

\vspace{-2mm}
\section{Killer Applications: The Needs for Deployment at the Edge\label{application}}
\vspace{-1mm}
LLMs can be directly applied or fine-tuned to a broad range of tasks. In this section, we will focus on two mission-critical use cases: healthcare and robotics control, to demonstrate the need for LLM deployment at the mobile edge.

\textbf{Healthcare:} Healthcare is widely recognized as a pivotal application for LLMs. Compared to traditional AI models, LLMs exhibit exceptional generalization, enhancing interactions with patients, caregivers, and medical professionals. \rev{Industry leaders have developed specialized LLMs, such as NVIDIA’s BioNeMo and Google’s Med-PaLM 2. Notably, Med-PaLM 2, an LLM fine-tuned on medical datasets, surpassed the pass mark on the US Medical License Exam (USMLE) and achieved 86.5\% accuracy.} Indeed, with multimodal inputs and outputs, LLMs can function as AI medical generalists, offering services from conversational support (e.g., chatbots) to diagnosis to early warnings. Nevertheless, the massive multimodal data transmissions may pose significant challenges for cloud-based healthcare LLM deployment. More importantly, cloud-based centralized training or inference faces substantial challenges in medical data collection owing to privacy concerns and data regulations, which necessitates privacy-preserving distributed learning paradigms, such as federated and split learning, to train/deploy models at the edge.


\textbf{Humanoid robots:} Robotic control is another critical application for LLMs. With remarkable generalization and reasoning capabilities, LLMs enable robots to comprehend human intention/emotion or complicated environments and plan sequential robotic manipulation accordingly. \rev{Several robotic LLMs have emerged, such as NVIDIA’s GR00T and Google's PALM-E~\cite{driess2023palm}. Notably, Google's PALM-E~\cite{driess2023palm}, adapted from a pre-trained LLM (i.e., PALM), can directly ingest raw streams of robotic sensor data, enabling robots to perform embodied reasoning and decompose a complex task (e.g., making a cake batter with ingredients the robot sees) into actionable steps.} Nevertheless, for robotics applications, centralized model training involves not only massive streaming video uploads, potentially overwhelming backhaul/backbone networks, but also sensitive interactive data relevant to daily human activities, in the form of voice and videos, leading to significant privacy threats. Moreover, since human-machine interactions and robotics maneuvers demand low-latency execution in various tasks (e.g., elderly/child care like preventing a kid from injury or poisoning), LLMs should be deployed at the network edge for facilitating real-time robotic control. 

Both of the applications underscore the importance of deploying LLMs at the network edge to address the \textit{bandwidth, latency, and privacy concerns}.

\vspace{-1mm}
\section{Limitations for On-device LLM Deployment\label{challenges}}
\vspace{-1mm}
As motivated above, there is a pressing need to deploy LLMs at the network edge. However, the current \textit{on-device} LLM deployment faces critical challenges. In this section, we identify these inherent limitations for on-device LLM deployment.



\textit{The first limitation arises from the extreme computational requirements.} 
\rev{The LLaMA-7B model, with 7 billion parameters, takes approximately 2.3 seconds to generate a single token on Apple M1 Max processor~\cite{alizadeh2024llm}. This latency is prohibitive for applications demanding real-time or long-form text generation, such as conversational AI and robotic decision-making. Without efficient optimization techniques, running LLMs at the edge may lead to unacceptably high latency.}

\textit{The second limitation stems from the intensive energy consumption.} \rev{A 7B-parameter LLM consumes 0.7 J per token~\cite{liu2024mobilellm}. For a fully charged iPhone with approximately 50kJ of energy, processing 700 tokens consumes around 1\% of the battery.} This necessitates offloading energy-consuming inference/fine-tuning tasks to edge servers to save battery power on mobile devices. On the other hand, from the perspective of our society, the overall energy consumption must be reduced to foster environmentally sustainable AI deployment.

\textit{Last but not least, storage and memory pose another limitation.} \rev{For instance, full-parameter fine-tuning of a 7-billion-parameter LLM with 16-bit precision requires approximately 84GB of memory, far exceeding the 8 GB RAM available on modern smartphones, such as the iPhone 16.} This memory requirement presents a significant obstacle in training LLMs.

\rev{Given the above limitations, instead of relying on on-device deployment, end-edge cooperation supported by modern 5G/6G MEC servers with high-performance GPUs (e.g., NVIDIA A100, H100), TPUs (e.g., Google Edge TPU), and AI accelerators (e.g., Ascend 910), may be a better deployment option for pushing LLMs to the network edge. In what follows, we will elaborate on end-edge cooperation for LLMs at the 6G edge.}

\begin{figure*}[t!]
\centering
\includegraphics[width=15.5cm]{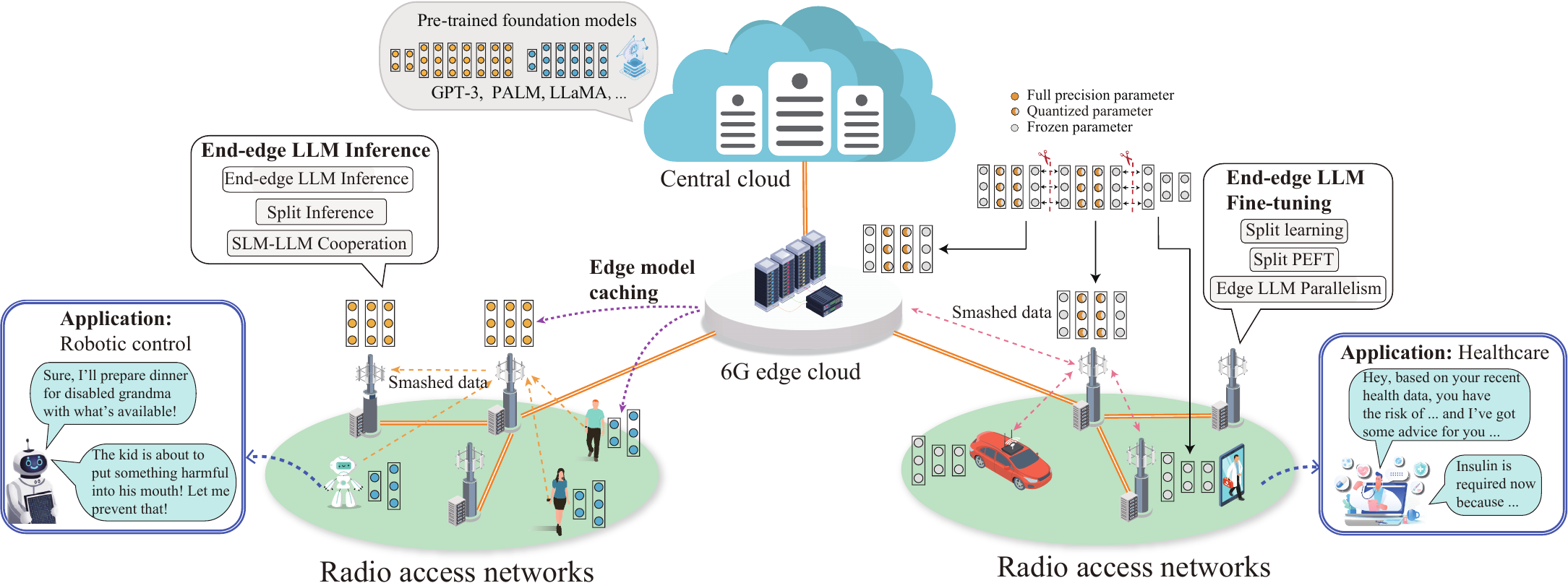}
\vspace{-0.2cm}
\caption{The MEC architecture for large language models in 6G.
}
\label{hierarchical_large_model}
\end{figure*}

\vspace{-2mm}
\section{6G MEC for Large Language Models: An Overview\label{sec:architecture}}
\vspace{-1mm}
As depicted in Fig \ref{hierarchical_large_model}, we introduce a 6G MEC architecture that fully supports the deployment of LLMs, consisting of the following critical modules.

\subsubsection{Cloud-edge-user synergy} 
Cloud servers, MEC servers, and users can cooperate in a \textit{hierarchical} manner. A central cloud retains a repository of pre-trained foundation models, such as the GPT series and PALM, applicable across various application scenarios. These models can be downloaded to edge networks and fine-tuned into downstream tasks with context-specific data. \rev{Several real-world edge AI platforms, such as NVIDIA Jetson AGX Orin and AWS Wavelength, have demonstrated the feasibility of deploying AI models at the edge.} For effective training and inference, cloud servers, edge servers, and users should collaborate in various manners. Specifically, split learning/inference can be implemented to optimize model splitting across cloud-edge-user networks to facilitate training/inference of an LLM, where splitting decisions are crucial for determining communication and computing workloads. Model splitting not only distributes the computational workload across resource-constrained devices and servers but also safeguards user privacy by eliminating raw data sharing. Moreover, cloud servers can deploy large-scale LLMs for accurate inference/training, whereas edge servers or edge devices can deploy small-sized LLMs with low training/inference latency, where the small and large models can cooperate with speculative decoding. \rev{Notably, our proposed MEC architecture aligns with the ongoing 6G standardization efforts by ITU~\cite{ITU-R_M.2160_2023}, particularly in AI-native networking, network function virtualization (NFV), and ultra-low-latency edge intelligence.} More details will be introduced in Section \ref{sec:fine_tuning} and Section \ref{inference}.

\subsubsection{In-network model splitting} 
Although the cloud possesses virtually unlimited resources, data transmission costs in the cloud-edge links may become the bottleneck. Apart from cloud-edge-user synergy, different MEC servers may also cooperate in a \textit{horizontal} manner, thus further reducing communication latency/costs. Within a network of MEC servers, models can be partitioned in different ways, i.e., layer-wise splitting or tensor splitting, for either inference or training. No matter how the model is partitioned, the intermediate smashed data (i.e., intermediate activations and back-propagated gradients), model parameters, or user data are exchanged across MEC servers. To harness distributed computing and storage resources for in-network model splitting, network virtualization is of paramount importance, which improves resource utilization, flexibility, and manageability. Following the design principle of software-defined networking, our envisioned 6G MEC architecture features a central controller that orchestrates network-wide computing resources and data transmissions, with the decoupled control and data plane. By collecting global network knowledge, the control partitions and coordinates model training/inference across the distributed edge computing systems.



\subsubsection{Parameter-sharing edge model caching and delivery\label{caching}} 
To support the functionalities above, the 6G MEC architecture must store, cache, and migrate LLMs in edge networks to enable fast model delivery for either downloading to users or distributed learning. In view of the size of LLMs, the strategic placement of LLMs on the appropriate edge servers must be carefully studied. Unlike traditional edge caching, 6G network operators can exploit the ``parameter sharing'' characteristics of LLMs to enable effective model placement and migration. Specifically, 
with parameter-efficient fine-tuning methods, such as LoRA, downstream LLMs can share most parameters from a pre-trained foundation model, with only a few trainable parameters for customization/personalization. This property can be exploited for storage-efficient model placement. Moreover, as users/servers request different models, only the task-specific parts of the models should be migrated to new sites with minimal model migration costs, thereby saving bandwidth.



\vspace{-2mm}
\section{End-edge Large Model Fine-tuning\label{sec:fine_tuning}}
\vspace{-1mm}
With LLMs pre-trained by the cloud, edge training can fine-tune them to new environments and personalize toward individual needs. In this section, we will elaborate on the end-edge co-training paradigm techniques for fine-tuning large language models (LLMs) at the network edge, focusing on split learning and edge LLM parallelism.

\begin{figure}[t]
\centering
\includegraphics[width=0.32\textwidth]{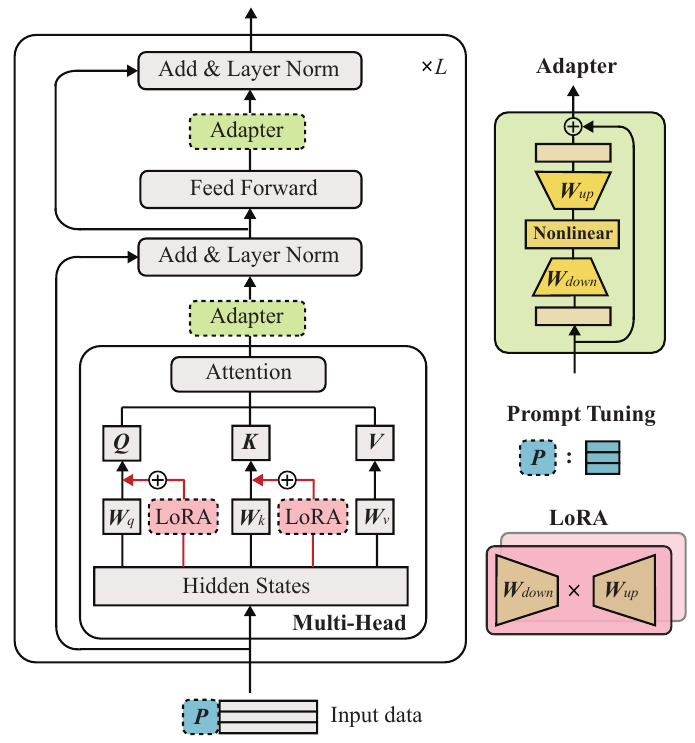}
\vspace{-0.25cm}
\caption{An illustration of the transformer architecture and several state-of-the-art PEFT methods, including adapter tuning, prompt tuning, and low-rank adaptation.
}
\label{finetune_method}
\end{figure}

\vspace{-2mm}
\subsection{Split Parameter-efficient Fine-tuning (SplitPEFT)\label{subsec:sl}}
\vspace{-1mm}
\subsubsection{Split learning} 
The limitations in computing power, memory, and storage on edge servers pose significant challenges for training LLM solely on a single edge device, leading to excessive latency and memory overflow. As a result, federated learning (FL) is typically unsuitable for LLM training. To address these issues, split learning (SL)~\cite{vepakomma2018split} has emerged as a promising solution. SL partitions a model into two sub-models and places them on clients and a server for collaborative training, where an edge device only trains several early layers, as illustrated in Fig.~\ref{hierarchical_large_model}. SL prevents raw data sharing with the server while significantly reducing the computing burden of local training. While the vanilla SL trains models between a server and clients in a sequential manner, later variants of SL, including parallel split learning (PSL) and split federated learning (SFL), parallelize the framework by enabling multiple devices to train a model with a server simultaneously, thereby further accelerating the process. A comprehensive review of the integration of SL and MEC systems, termed split edge learning (SEL),
can be found in~\cite{lin2023split}.

\subsubsection{Integrating SL and parameter-efficient fine-tuning}
SL should be combined with the state-of-the-art LLM fine-tuning techniques to support effective end-edge co-training at the network edge. To fine-tune LLMs on edge devices/servers, conventional full-parameter fine-tuning (i.e., updating all parameters) is prohibitively expensive, even when the workload is shared between edge devices and an edge server. Besides, for SFL, full-parameter fine-tuning also incurs considerable communication costs associated with model aggregation. 

To address the above concerns, parameter-efficient fine-tuning (PEFT) techniques are often necessary. Specifically, PEFT updates only a tiny fraction of model parameters, to adapt LLMs effectively to new tasks or environments, significantly reducing training and communication overhead while mitigating overfitting. As shown in Fig.~\ref{finetune_method}, there are several representative PEFT approaches for LLMs, including adapter tuning, prompt tuning, and Low-rank adaptation (LoRA). Adapter tuning involves inserting well-designed adapter modules between layers for training, while prompt tuning adds tunable prefix tokens. LoRA decomposes the weights of attention layers into low-rank matrices for updating without adding additional inference costs. The shared principle among these methods is to train a small number of parameters, typically less than 1\% of the original parameters, dramatically decreasing the number of trainable parameters, thereby reducing training latency, communication latency, and memory consumption.

\begin{figure}[t]
\centering
\includegraphics[width=0.23\textwidth]{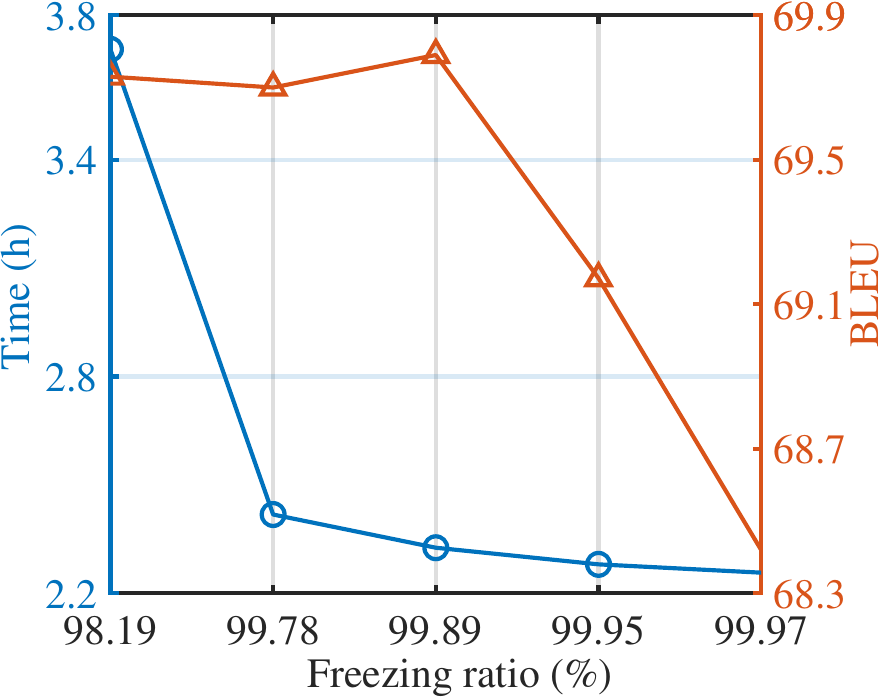}
\vspace{-0.25cm}
\caption{The performance of SplitLoRA~\cite{lin2024splitlora} for training latency and bilingual evaluation understudy (BLEU, a metric for evaluating the machine translations against the human translations) versus the freezing ratio, where LoRA is employed to fine-tune GPT-2 medium on WebText dataset. An edge server and $20$ clients are considered. Computing capabilities of clients and the edge server are set to 3.56 and 35.6 (peak performance of one NVIDIA RTX 3090) TFLOPS, uplink and downlink rates are 70Mbps and 300Mbps, and the number of tokens utilized for training is 264M.
}
\label{finetune_latency_acc}
\end{figure}

By integrating SL and PEFT, our work~\cite{lin2024splitlora} proposes SplitLoRA, the first framework to integrate SL and LoRA PEFT. SplitLoRA combines the strengths of SL - offloading substantial training workloads to the server via model partitioning - and LoRA - updating a small portion of model parameters to enable efficient LLM adaptation. As shown in Fig.~\ref{finetune_latency_acc},  we observe model training can be accomplished in a reasonable time on the edge server with an NVIDIA RTX 3090. \rev{However, this efficiency comes with a trade-off: while a higher freezing ratio significantly decreases communication-computing latency, it may compromise model performance due to restricted parameter updates.} \rev{Future research should focus on striking the optimal trade-off between accuracy and latency by controlling the optimal ratios in SplitLoRA. For instance, server-side model training can be conducted with a lower freezing ratio to preserve model accuracy, whereas client-side training can be frozen more to reduce client-side computing overhead.} How to jointly determine the optimal freezing ratio is an interesting research direction.

\subsubsection{Towards efficient split PEFT}
Although Split PEFT provides a feasible solution for end-edge fine-tuning, the computing-communication workload is still intensive. For instance, if considering the splitting of LLMs for end-edge co-training/co-inference, the total smashed data volume at the cut layer can be approximately 420 MB for one training round if considering GPT-3 and 100 data samples of 1024 tokens. Besides, the fine-tuning workload can still be prohibitive for edge devices.

To design a more efficient split PEFT, research efforts can be made in different aspects: 1) The optimal model split strategies can be investigated in combination with freezing ratios to balance the communication and computing workload. Selecting the cut layer is a fundamental problem in SFL, which affects not only communication-computing overhead but also model convergence. In split PEFT, this tradeoff becomes more complex because it also impacts the freezing ratios on both the device and server sides. For example, selecting a shallow cut layer (fewer layers on the client side) may allow clients to train models with more unfrozen parameters. As a result, these two decision variables, i.e., model splitting and freezing ratios, are tightly coupled in split PEFT. 2) Client-side models can be entirely frozen. In this case, clients can only perform one forward pass. More importantly, since the smashed data output from the frozen client-side models will not be changed across different training epochs, the smashed data upload needs to be carried out in the \textit{first} training epoch only, significantly saving communication bandwidth. However, in this case, the model splitting decision should be carefully carried out to ensure that the server-side model training is sufficient for training performance. 3) Smashed data compression and error control should be conducted to reduce the uploading data volume without noticeably affecting training accuracy. In general, model training is robust to lossy transmissions. For this reason, joint smashed data compression and error corrections, analogous to joint source-channel coding, can be explored to enhance the transmission efficiency of SL.

\begin{figure}[t]
\centering
\includegraphics[width=0.40\textwidth]{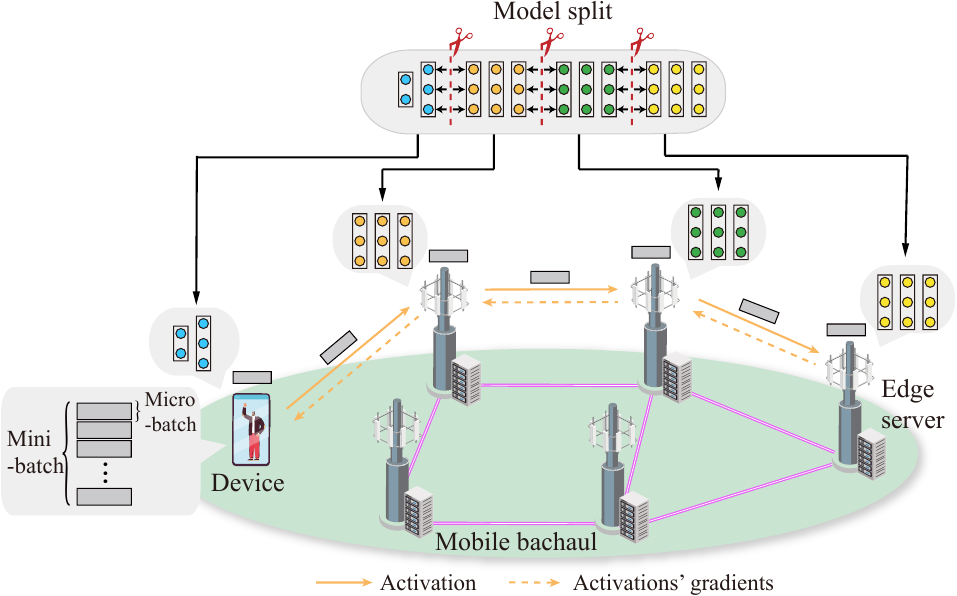}
\vspace{-0.25cm}
\caption{An illustration of multi-hop SL with pipeline parallelism. Multiple clients jointly train a large model based on SL approaches, such as SFL and PSL. The model is partitioned into multiple parts so that the total workload is shared among multiple edge servers.
}
\label{multi_hop_SL_system}
\end{figure}

\vspace{-3mm}
\subsection{Parallelism of Edge LLM Training\label{subsec:sl}}
\vspace{-1mm}
Considering the size of LLMs, splitting a model into two submodels might still be insufficient. To address this issue, the 6G edge can partition an LLM into more partitions based on various parallelism strategies, including pipeline parallelism, tensor parallelism, and data parallelism, placing them on different MEC servers within edge networks. Among these approaches, pipeline parallelism can be considered as multi-hop SL, enabling multiple edge servers to work collaboratively to train the model, where a training mini-batch into multiple micro-batches for parallel processing across edge servers in a multi-hop manner, as illustrated in Fig. \ref{multi_hop_SL_system}; Tensor parallelism distributes individual operators in a layer across different processors/servers; Data parallelism allows for model replication across various processors/servers with training datasets. To accommodate for large-scale LLM training, Narayanan et al. \cite{narayanan2021efficient} combined tensor, pipeline, and data parallelism to efficiently train LLMs by harnessing distributed GPUs in cloud centers. These methods can also be utilized to support LLM training at the network edge.

\begin{figure}[t]
\centering
\includegraphics[width=0.23\textwidth]{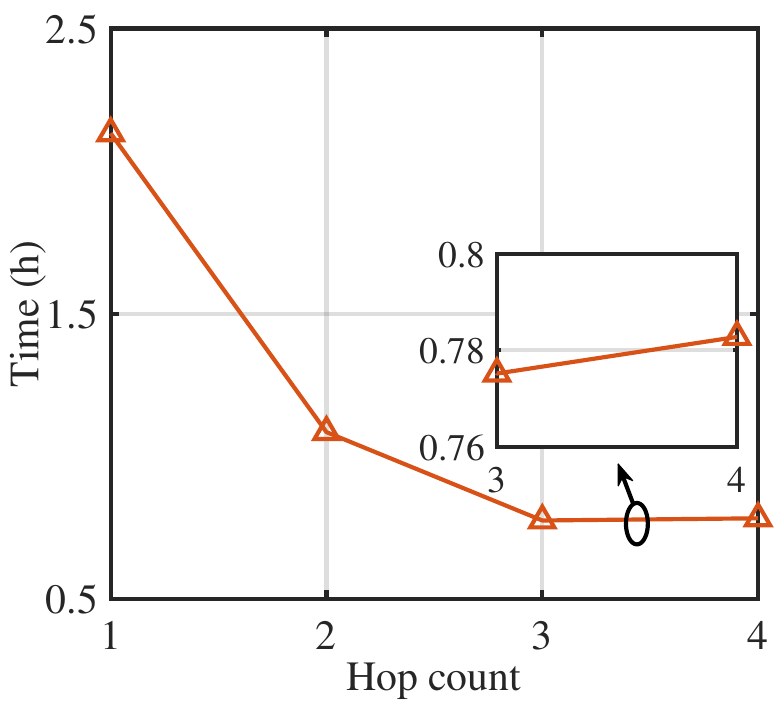}
\vspace{-0.25cm}
\caption{The training latency of multi-hop SL versus the hop counts, where LoRA is employed to fine-tune GPT-2 medium on WebText dataset. The data samples are distributed over $5$ clients, the transmission rate between edge servers is 400Mbps, and other key parameters are consistent with Fig. \ref{finetune_latency_acc}.
}
\label{multi_hop_SL_latency}
\end{figure}

A crucial distinction between LLM parallelism at the network edge and in GPU clusters, such as in \cite{narayanan2021efficient}, is \textit{resource heterogeneity}. Since model training in edge networks involves servers with significantly heterogeneous computing, memory, and communication capabilities, inappropriate model splitting and placement can result in bottleneck servers/links, leading to intolerable latency. Instead, distributed training in GPU clusters which normally assumes identical GPUs and communication links. Considering wireless edge networks, Fig. \ref{multi_hop_SL_latency} demonstrates the total training time (including both computing and communication latency) for achieving a target accuracy for the LLM GPT-2 medium. Although leveraging more edge servers takes full advantage of distributed computing resources, it increases communication overhead and potentially longer end-to-end latency. For this reason, jointly optimizing model splitting and placement under heterogeneous resource constraints is vital in enhancing the efficiency of LLM parallelism at the network edge. As a result, more research can be done by various parallelisms, i.e., model, data, and tensor parallelism strategies, for diserpsing large-scale LLM training in resource-constrained edge networks.


\vspace{-2mm}
\section{End-edge Large Model Inference\label{inference}}
\vspace{-1mm}
Model inference refers to running input data into a model to get the outputs. Cloud-based AI model inference incurs significant communication latency, which violates the service requirements of many applications as motivated in Section \ref{challenges}. Nevertheless, despite low-latency data transmissions, edge devices usually possess limited computing resources, which might incur long computing latency. In this section, we address the challenge by presenting end-edge co-inference at the network edge.
\par


\vspace{-4mm}
\subsection{Split inference for LLMs}
\vspace{-1mm}
Split inference is a model inference technique that offloads the computing workload from edge devices to a server via model partitioning. At the 6G edge networks, this technique enables edge devices to cooperate with an MEC server for efficient inference.  Beyond enhancing privacy preservation, split inference can reduce communication overhead if the cut-layer feature size is smaller than raw data size. This advantage is particularly pronounced for multimodal LLMs that use high-definition images or videos as input data. While split inference has been studied in the past years, there are some new characteristics of design for LLMs, as detailed below.

\subsubsection{Split inference with KV caching} The state-of-the-art LLMs are mostly autoregressive LLMs, implying they generate sequences by predicting each token sequentially based on the previous tokens. When employing split inference for autoregressive LLMs, the volume of smashed data exchange at the cut layer may increase with the length of input tokens (as more generated output tokens will be packed into the input tokens), resulting in a communication bottleneck.

To tackle this challenge, model splitting can be implemented with KV (key-value) caching~\cite{kang2024gear}, as widely adopted in autoregressive LLMs to store previous Key and Values for the preceding tokens. Consequently, the size of smashed data exchanged between edge devices and servers only pertains to the newly added token rather than all previous tokens. However, KV caching trades memory consumption for less communication and computing workload. When memory is limited on edge devices and servers, KV caching may not be effective for long input sequences and a large number of served users due to excessive memory requirements. Since model splitting impacts memory, computing, and communication load between edge devices and servers, joint optimization of model splitting and KV caching can be explored by considering these resource constraints, including memory constraints.


\subsubsection{Parameter-sharing split inference} To support split inference in multi-user systems, an edge device/server may run multiple LLM-empowered applications simultaneously, consuming significant running memory. For example, a single inference task for FP16 GPT 6.7B with 512 input sequences and 32 output sequences approximately requires 41.84 GB of running memory when the batch size is set as 64 \cite{nvidia2023faster}. As the number of running applications increases and the sequence length grows, the running memory requirement will be a major bottleneck for an edge server. Moreover, for LLMs with billions of parameters, e.g., GPT-2, model loading to GPU memory becomes dominant in end-to-end latency. Thankfully, as discussed in Section \ref{caching}, parameter sharing is a prevalent feature among LLMs. By exploiting this property, when multiple models comprise the same parameters or blocks, an edge server can load only a single copy of the shared parameters into the GPU memory, thereby substantially reducing memory costs and data swapping time. To this aim, split inference for multi-user user scheduling can be optimized by exploiting the shared parameters among LLMs on an edge server.

\subsubsection{Split inference with Mixture of Experts (SplitMoE)}
To further enable the computing efficiency of LLM, split inference can be combined with mixture-of-experts (MoE). \rev{The MoE architecture~\cite{wang2024toward} is an efficient and scalable neural network paradigm that divides a model into multiple specialized expert sub-networks, with a gating mechanism dynamically determining the most relevant experts for each input, enabling effective input processing while significantly reducing both computational and memory overheads.} Split inference partitions model execution between edge devices and servers, while MoE assigns specialized experts to handle diverse tasks, enhancing computational efficiency. The combination of split inference and MoE offers a transformative solution for deploying large-scale models in resource-constrained environments, where an edge device can offload its workload to specific edge servers with the most appropriate experts. \rev{However, dynamic expert selection and placement across distributed edge networks introduce new challenges, such as optimizing expert assignment to balance accuracy, latency, and network load. This synergy between MoE and split inference not only improves resource utilization but also ensures superior performance in latency-sensitive applications.} An interesting direction is how to jointly optimize expert partitioning and scheduling in edge networks for effective split inference and training.

\vspace{-3mm}
\subsection{Small-Large Language Model (SLM-LLM) Cooperation} 
\vspace{-1mm}
Although split inference reduces client-side workload significantly, the intermediate features at the cut layer are often high-dimensional, resulting in long communication latency. To mitigate this issue, end-edge co-inference can also be executed in the following way: edge devices execute small language models (SLM) for fast execution, whereas an edge server runs LLM occasionally to verify/correct the results from SLM. Since the output tokens from SLM are often very small, the communication overhead can be reduced substantially.

\subsubsection{SLM-LLM cooperation with speculative decoding}
The SLM-LLM cooperation can be naturally combined with the state-of-the-art speculative decoding process of LLMs to reduce inference latency significantly~\cite{leviathan2023fast}. Specifically, LLMs typically generate tokens autoregressively, which requires running the model K times to produce K tokens, resulting in significant latency. Speculative decoding addresses this issue by using a smaller model to generate multiple tokens, which are verified and corrected by the larger model in parallel. This method is much faster than the standard autoregressive generation in LLMs, as it eliminates the autoregressive process on the server side. Based on this idea, a hybrid collaborative approach~\cite{hao2024hybrid} has been proposed, where SLMs are deployed on edge devices to generate draft tokens, which are then verified and corrected by an LLM on the server based on prediction confidence and acceptance threshold. Experiments on datasets like GSM8K demonstrate that such methods can achieve LLM-level performance with only about 25.8\% of computing cost. However, this work does not adaptively control the thresholds according to communication-computing resources. To further enhance SLM-LLM in wireless networks, an adaptive protocol should be designed to adjust the acceptance threshold or the length of draft tokens based on the output confidence of SLMs, available computing capabilities, and wireless channel conditions, which is worth exploring in the future.

\vspace{-3mm}
\section{Open problems\label{Open}}
\vspace{-1mm}
As an emerging field, there are still numerous open research problems on how to employ MEC systems to support LLMs, particularly on the end-edge cooperation aspect. We pick up a few most important ones to discuss as follows.

\vspace{-3mm}
\subsection{Green and Sustainable End-edge Cooperation}
\vspace{-1mm}
Despite their significantly powerful capabilities, training and
inference of LLMs are notoriously power-hungry due to their
huge size. \rev{To minimize energy consumption while maintaining satisfactory model performance, MEC systems must intelligently schedule model training, carefully select high-quality data for training, and judiciously determine which model to use. To reduce the overall energy footprint, MEC systems can run smaller LLMs for less complex tasks, potentially on devices, while executing
large-sized models on the edge server only for challenging
tasks. The training and inference efficiency can be further improved by developing uncertainty-aware decision mechanisms, enabling SLMs to determine when to rely on LLM corrections based on confidence scores. To save energy on mobile devices, a key strategy is the joint optimization of model splitting and PEFT, where adaptive freezing ratios can dynamically shift more energy demands to the fixed infrastructure under the same accuracy constraints. Future research should explore reinforcement learning-based or gradient-based strategies to determine optimal freezing ratios, balancing energy efficiency, communication overhead, and training accuracy. }All of these require innovative network optimization for energy-efficient LLM training and inference at the mobile edge.


\vspace{-4mm}
\subsection{Privacy-preserving End-edge Cooperation}
\vspace{-1mm}
While both SL and SFL can enhance privacy for LLM, it has been demonstrated that smashed data and model parameters might still pose risks of privacy breaches for data owners. For LLM applications, this challenge becomes prominent because it has been shown that prompting the LLMs can extract the private information, such as credit card information, of other users participating in the training process by only inserting a few benign-appearing sentences into the training dataset. \rev{Similarly, in SLM-LLM cooperation, transmitting draft tokens from the SLM to the LLM may expose sensitive user input. To offer more robust privacy protection, noise injection techniques such as differential privacy and homomorphic encryption can be employed, particularly for highly sensitive training data and draft tokens, while still preserving the model performance.} For edge-device cooperation in wireless networks, it has been shown that such privacy preservation can naturally be incorporated by using wireless channel noises, thus turning a negative aspect into a beneficial one. This requires the co-design of privacy-preserving mechanisms and wireless communications under the LLM context.

\vspace{-3mm}
\section{Conclusions\label{conclusion}}
\vspace{-2mm}
In recent years, language models have experienced exponential growth in size, giving birth to numerous LLMs with billions of parameters. This trend urges us to think about how edge intelligence can accommodate these giant models. In this article, we advocated the paradigm shift from cloud computing to 6G MEC for LLM deployment with a focus on end-edge cooperation. We highlighted killer applications to motivate this paradigm shift from the cloud to the edge. Meanwhile, we identified the key challenges that mainly arise from the on-device LLMs. To address these challenges, we first proposed a 6G MEC architecture for LLMs and then elaborated on enabling techniques for efficient end-edge LLM fine-tuning and inference. We hope this article can inspire more researchers in the wireless community to explore the deployment of LLMs at the mobile edge and further advance this emerging field.
\vspace{-5mm}
\bibliographystyle{IEEEtran}
\bibliography{NEWmybib}

\begin{thebibliography}{10}
\providecommand{\url}[1]{#1}
\csname url@samestyle\endcsname
\providecommand{\newblock}{\relax}
\providecommand{\bibinfo}[2]{#2}
\providecommand{\BIBentrySTDinterwordspacing}{\spaceskip=0pt\relax}
\providecommand{\BIBentryALTinterwordstretchfactor}{4}
\providecommand{\BIBentryALTinterwordspacing}{\spaceskip=\fontdimen2\font plus
\BIBentryALTinterwordstretchfactor\fontdimen3\font minus \fontdimen4\font\relax}
\providecommand{\BIBforeignlanguage}[2]{{%
\expandafter\ifx\csname l@#1\endcsname\relax
\typeout{** WARNING: IEEEtran.bst: No hyphenation pattern has been}%
\typeout{** loaded for the language `#1'. Using the pattern for}%
\typeout{** the default language instead.}%
\else
\language=\csname l@#1\endcsname
\fi
#2}}
\providecommand{\BIBdecl}{\relax}
\BIBdecl

\bibitem{achiam2023gpt}
J.~Achiam, S.~Adler, S.~Agarwal, L.~Ahmad, I.~Akkaya, F.~L. Aleman, D.~Almeida, J.~Altenschmidt, S.~Altman, S.~Anadkat \emph{et~al.}, ``{Gpt-4 Technical Report},'' \emph{arXiv preprint arXiv:2303.08774}, Mar. 2023.

\bibitem{shen2023large}
Y.~Shen, J.~Shao, X.~Zhang, Z.~Lin, H.~Pan, D.~Li, J.~Zhang, and K.~B. Letaief, ``{Large Language Models Empowered Autonomous Edge {AI} for Connected Intelligence},'' \emph{{IEEE} Commun. Mag.}, Jan. 2024.

\bibitem{driess2023palm}
D.~Driess, F.~Xia, M.~S. Sajjadi, C.~Lynch, A.~Chowdhery, B.~Ichter, A.~Wahid, J.~Tompson, Q.~Vuong, T.~Yu \emph{et~al.}, ``{Palm-e: An Embodied Multimodal Language Model},'' \emph{Proc. ICML}, Jul. 2023.

\bibitem{alizadeh2024llm}
K.~Alizadeh, S.~I. Mirzadeh, D.~Belenko, S.~Khatamifard, M.~Cho, C.~C. Del~Mundo, M.~Rastegari, and M.~Farajtabar, ``{LLM in a Flash: Efficient Large Language Model Inference with Limited Memory},'' in \emph{{Proc. ACL}}, Jul. 2024.

\bibitem{liu2024mobilellm}
Z.~Liu, C.~Zhao, F.~Iandola, C.~Lai, Y.~Tian, I.~Fedorov, Y.~Xiong, E.~Chang, Y.~Shi, R.~Krishnamoorthi \emph{et~al.}, ``{MobileLLM: Optimizing Sub-billion Parameter Language Models for On-Device Use Cases},'' \emph{arXiv preprint arXiv:2402.14905}, Feb. 2024.

\bibitem{ITU-R_M.2160_2023}
{ITU-R}, ``{IMT-2030 Framework and Overall Objectives of the Future Development of IMT for 2030 and Beyond},'' \emph{https://www.itu.int/rec/R-REC-M.2160}, 2023.

\bibitem{vepakomma2018split}
P.~Vepakomma, O.~Gupta, T.~Swedish, and R.~Raskar, ``{Split Learning for Health: Distributed Deep Learning Without Sharing Raw Patient Data},'' \emph{arXiv preprint arXiv:1812.00564}, Dec. 2018.

\bibitem{lin2023split}
Z.~Lin, G.~Qu, X.~Chen, and K.~Huang, ``{Split Learning in 6G Edge Networks},'' \emph{{IEEE} Wireless Commun.}, Jan. 2024.

\bibitem{lin2024splitlora}
Z.~Lin, X.~Hu, Y.~Zhang, Z.~Chen, Z.~Fang, X.~Chen, A.~Li, P.~Vepakomma, and Y.~Gao, ``{Splitlora: A Split Parameter-efficient Fine-tuning Framework for Large Language Models},'' \emph{arXiv preprint arXiv:2407.00952}, Jul. 2024.

\bibitem{narayanan2021efficient}
D.~Narayanan, M.~Shoeybi, J.~Casper, P.~LeGresley, M.~Patwary, V.~Korthikanti, D.~Vainbrand, P.~Kashinkunti, J.~Bernauer, B.~Catanzaro \emph{et~al.}, ``{Efficient Large-scale Language Model Training on GPU Clusters Using Megatron-LM},'' in \emph{Proc. SC}, Nov. 2021, pp. 1--15.

\bibitem{kang2024gear}
H.~Kang, Q.~Zhang, S.~Kundu, G.~Jeong, Z.~Liu, T.~Krishna, and T.~Zhao, ``{Gear: An Efficient KV Cache Compression Recipefor Near-lossless Generative Inference of LLM},'' \emph{arXiv preprint arXiv:2403.05527}, Mar. 2024.

\bibitem{nvidia2023faster}
\BIBentryALTinterwordspacing
Nvidia, ``Fastertransformer,'' 2023. [Online]. Available: \url{https://github.com/NVIDIA/FasterTransformer}
\BIBentrySTDinterwordspacing

\bibitem{wang2024toward}
J.~Wang, H.~Du, D.~Niyato, J.~Kang, Z.~Xiong, D.~I. Kim, and K.~B. Letaief, ``{Toward Scalable Generative AI via Mixture of Experts in Mobile Edge Networks},'' \emph{arXiv preprint arXiv:2402.06942}, Feb. 2024.

\bibitem{leviathan2023fast}
Y.~Leviathan, M.~Kalman, and Y.~Matias, ``{Fast Inference From Transformers via Speculative Decoding},'' in \emph{Proc. ICML}, 2023.

\bibitem{hao2024hybrid}
Z.~Hao, H.~Jiang, S.~Jiang, J.~Ren, and T.~Cao, ``{Hybrid SLM and LLM for edge-Cloud Collaborative Inference},'' in \emph{Proceedings of the Workshop on Edge and Mobile Foundation Models}, 2024, pp. 36--41.

\end{thebibliography}
\vspace{-1cm}

\begin{IEEEbiographynophoto}{Zheng Lin} is currently pursuing his Ph.D. degree at the University of Hong Kong.  His research interests include edge intelligence and distributed machine learning.
\end{IEEEbiographynophoto}

\vspace{-1.1cm}

\begin{IEEEbiographynophoto}{Guanqiao Qu} is currently pursuing his Ph.D. degree at the University of Hong Kong. His research interests include edge intelligence and federated learning.
\end{IEEEbiographynophoto}

\vspace{-1.1cm}
\begin{IEEEbiographynophoto}{Qiyuan Chen} is currently pursuing his Ph.D. degree at the University of Hong Kong. His research interests include edge intelligence and distributed learning.\end{IEEEbiographynophoto}

\vspace{-1.1cm}
\begin{IEEEbiographynophoto}{Xianhao Chen} is an assistant professor at the Department of Electrical and Electronic Engineering, The University of Hong Kong. His research interests include wireless networking, edge intelligence, and distributed learning.
\end{IEEEbiographynophoto}

\vspace{-1.1cm}
\begin{IEEEbiographynophoto}{Zhe Chen} is an assistant professor with the School of Computer Science at Fudan University. His research interests include wireless networking, deep learning, mobile and pervasive computing, and embedded systems.
\end{IEEEbiographynophoto}

\vspace{-1.1cm}
\begin{IEEEbiographynophoto}{Kaibin Huang} [Fellow, IEEE] is a Professor at the Department of Electrical and Electronic Engineering, The University of Hong Kong. His research interests include mobile edge computing, edge AI, and 6G systems.
\end{IEEEbiographynophoto}
\end{document}